\documentclass[letterpaper]{article} %
\usepackage[preprint]{aaai2027}  %
\usepackage[hyphens]{url}  %
\usepackage{graphicx} %
\usepackage{natbib}  %
\usepackage{caption} %
\usepackage{algorithm}
\usepackage{algorithmic}

\usepackage{newfloat}
\usepackage{listings}
\DeclareCaptionStyle{ruled}{labelfont=normalfont,labelsep=colon,strut=off} %
\floatstyle{ruled}
\newfloat{listing}{tb}{lst}{}
\floatname{listing}{Listing}

\usepackage{booktabs}
\usepackage{multirow}
\usepackage{tabularx}
\usepackage{amsmath,amssymb}
\newcolumntype{Y}{>{\centering\arraybackslash}X}

\usepackage{xcolor}
\usepackage{colortbl}
\newcommand{\method}{Relational Geometry Forcing}
\newcommand{\methodabbr}{RGF}
\newcommand{\model}{RGF-WAM}

\title{\method{} for World Action Models}
\author{
    \normalsize
    Yuan Xu\textsuperscript{1,2}\textsuperscript{*}\quad
    Yixiang Chen\textsuperscript{1,2}\quad
    Qisen Ma\textsuperscript{1,2}\quad
    Jiabing Yang\textsuperscript{1,2}\quad
    Peiyan Li\textsuperscript{1,2} \\[2pt]
    \normalsize
    Kai Wang\textsuperscript{1,2}\quad
    Jianhua Yang\textsuperscript{1,2}\quad
    Jianlou Si\textsuperscript{3}\quad
    Jun Huang\textsuperscript{3} \\[2pt]
    \normalsize
    Jing Liu\textsuperscript{1,2,4}\quad
    Nianfeng Liu\textsuperscript{1,2,4}\quad
    Yan Huang\textsuperscript{1,2,4}\quad
    Liang Wang\textsuperscript{1,2}
}
\affiliations{
    \textsuperscript{1}CASIA\quad
    \textsuperscript{2}UCAS\quad
    \textsuperscript{3}Alibaba Group\quad
    \textsuperscript{4}FiveAges
}

\begin{document}

\maketitle
\begingroup
\renewcommand{\thefootnote}{\fnsymbol{footnote}}
\footnotetext[1]{\texttt{yuan.xu@nlpr.ia.ac.cn}}
\endgroup

\begin{abstract}
World Action Models (WAMs) have shown strong potential for robot manipulation by jointly learning visual dynamics and executable actions.
However, their video and action objectives do not explicitly constrain the 3D geometry that is crucial for precise manipulation.
Recently, geometry-aware WAMs introduce depth or 4D prediction branches, whereas representation alignment methods supervise intermediate WAM features with pretrained 3D foundation models.
Yet we find that such pointwise alignment benefits the video features but degrades action learning, since dense geometric feature targets overly restrict compact action features.
To address this limitation, we propose \method{} (\methodabbr{}), which transfers geometric priors from a frozen 3D foundation model by aligning pairwise feature relations instead of pointwise features.
Specifically, \methodabbr{} applies three complementary relational objectives, where intra-frame and cross-view relations transfer scene geometry and multi-view correspondence to the video features, while action-temporal relations couple action dynamics to geometric scene evolution.
Applied only during training, \methodabbr{} requires no explicit geometry branch or extra inference module.
Extensive experiments in simulation and the real world demonstrate the effectiveness of \method{}, showing consistent gains across diverse manipulation scenarios and WAM paradigms.
\end{abstract}

\section{Introduction}
\label{sec:introduction}

\begin{figure}[!t]
    \centering
    \includegraphics[width=\columnwidth]{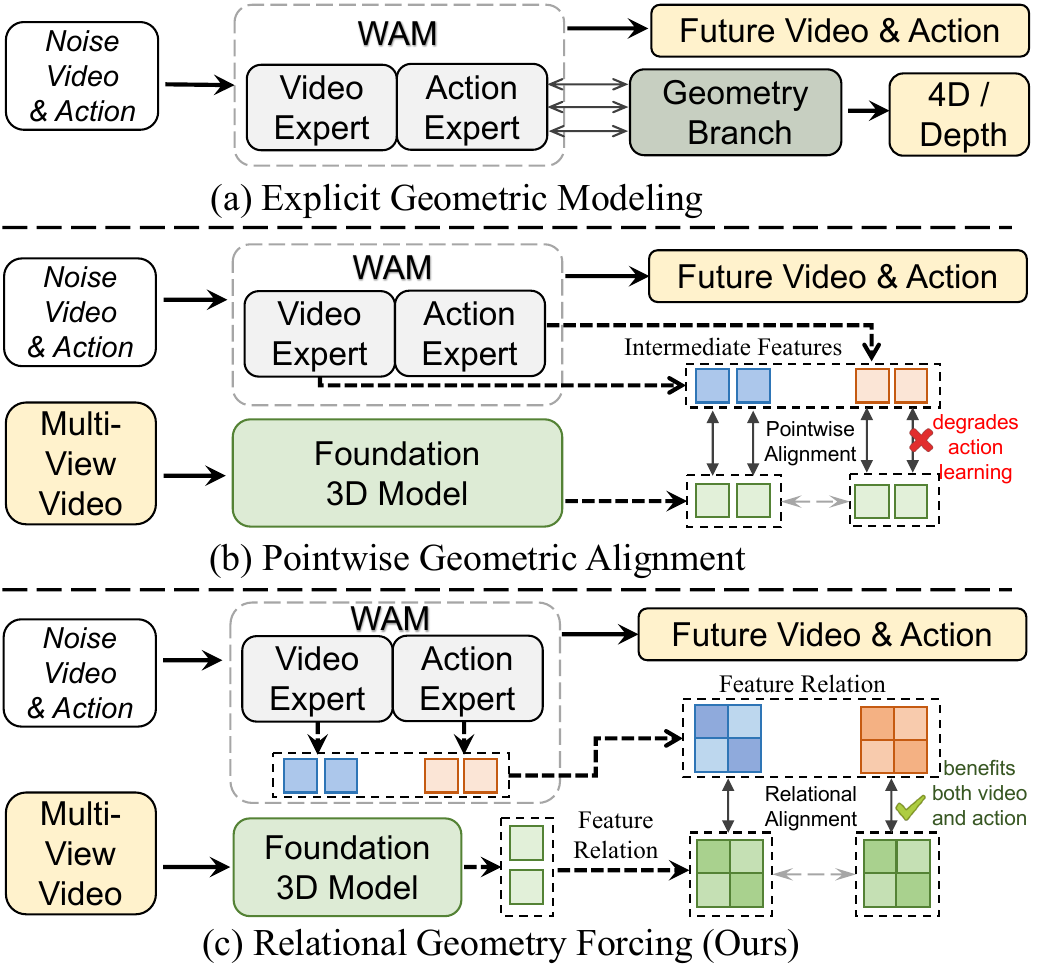}
    \caption{\textbf{Geometry transfer paradigms for WAMs.}
    (a) Explicit geometric modeling adds a geometry branch that predicts depth or 4D outputs.
    (b) Pointwise geometric alignment regresses intermediate features toward dense geometric targets, which benefits the video expert but degrades action learning.
    (c) \methodabbr{} aligns pairwise feature relations instead, benefiting both the video and action experts.}
    \label{fig:teaser}
\end{figure}

Robot policies for manipulation require accounting for interaction-induced scene changes when generating precise action trajectories.
World Action Models (WAMs) incorporate predictive world modeling into action learning by coupling future visual prediction with action generation~\cite{dreamzero,lingbotva,cosmospolicy,bi2026motus}.
Typically, WAMs include a video expert to predict future RGB frames and an action expert to generate action chunks, while scene geometry receives no direct supervision~\cite{yuan2026fastwam,gigaworldpolicy}.
Accordingly, limited geometric perception of the physical scene remains a key challenge for WAMs in precise manipulation~\cite{shridhar2022peract,ze2024dp3}.

Recent geometry-aware WAMs address this challenge by adding explicit geometry branches to the video--action framework, as illustrated in Figure~\ref{fig:teaser}(a).
These branches introduce depth or 4D prediction as additional training objectives~\cite{guo2026xwam,ma2026geosemwam,li2026wam4d,zhang2026mecowam}.
We therefore ask whether geometric priors can directly regularize the video and action representations of WAMs without explicit geometric reconstruction.

Representation alignment offers an alternative to explicit geometric modeling by directly supervising intermediate WAM features.
REPA aligns diffusion features with pretrained visual representations~\cite{yu2025repa}, while subsequent methods use geometric targets from pretrained 3D foundation models to supervise video generators and robot policies~\cite{wu2026geometry,li2026spatial}.
As illustrated in Figure~\ref{fig:teaser}(b), these methods primarily use pointwise alignment to regress intermediate features in WAMs toward dense geometric targets from 3D foundation models.

\begin{figure}[t]
    \centering
    \includegraphics[width=\columnwidth]{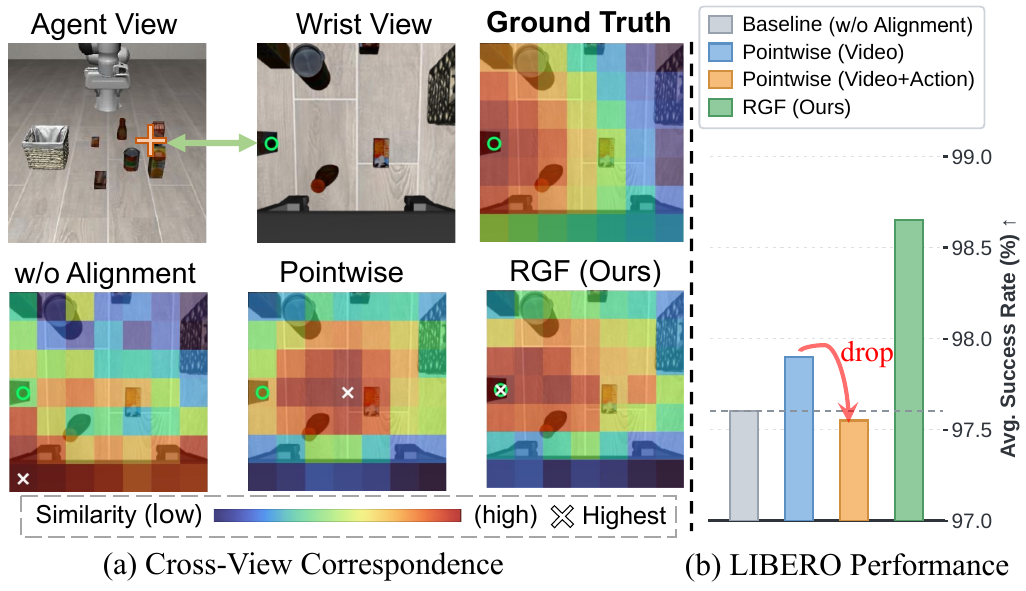}
    \caption{\textbf{Cross-view correspondence and LIBERO performance.}
    (a) Wrist-view similarity maps for a query patch in the agent view, where \methodabbr{} peaks near the ground-truth correspondence.
    (b) Comparison of average success rate on the same WAM: extending pointwise alignment to action features degrades performance below the baseline.}
    \label{fig:teaser-result}
\end{figure}

However, pointwise alignment assumes that each WAM feature and its geometric target encode comparable content.
Video alignment satisfies this assumption because both representations describe local image regions.
In action alignment, compact action features instead summarize trajectory transitions, whereas dense geometric targets describe scene content.
Directly regressing action features toward scene-level geometric targets therefore conflates representations with different roles.
Consistent with this analysis, extending pointwise alignment to the action expert reduces task success below the unaligned baseline (Figure~\ref{fig:teaser-result}b).

To address this problem, we introduce \method{} (\methodabbr{}), which transfers geometric priors to WAM features without explicit geometric reconstruction.
As illustrated in Figure~\ref{fig:teaser}(c), \methodabbr{} aligns pairwise relations among WAM features with those induced by a frozen 3D foundation model, avoiding direct feature regression.
Matching these relations does not require WAM and 3D features to share the same dimensionality or embedding space~\cite{park2019rkd,zhang2025videorepa}.
Using frozen VGGT features as supervision~\cite{wang2025vggt}, \methodabbr{} transfers scene geometry and multi-view correspondence to video features through intra-frame and cross-view objectives, while its action-temporal objective couples action dynamics to geometric scene evolution.
All auxiliary components are used only during training, leaving inference unchanged.

Experiments on LIBERO~\cite{liu2023libero}, RoboTwin 2.0~\cite{chen2025robotwin}, and real-world tasks on a Franka robot show that \methodabbr{} achieves competitive performance across single-arm and bimanual manipulation.
On the real robot, \methodabbr{} further generalizes to unseen objects, background distractors, and height changes.
Moreover, \methodabbr{} yields consistent gains across diverse WAM paradigms.
Qualitatively, Figure~\ref{fig:teaser-result}(a) shows that the \methodabbr{} similarity map peaks near the ground-truth cross-view correspondence.

Our contributions are as follows:
\begin{itemize}
    \item We identify a representation mismatch in pointwise alignment between dense geometric targets from 3D foundation models and compact action features in WAMs, explaining the performance degradation observed when this alignment is extended to the action expert.
    \item We propose \method{}, which transfers geometric priors into WAM representations by aligning intra-frame, cross-view, and action-temporal feature relations, without explicit geometric reconstruction or additional inference modules.
    \item Extensive experiments on simulated and real-world tasks validate \methodabbr{}, demonstrating competitive performance, generalization to unseen real-world conditions, and consistent gains across WAM paradigms.
\end{itemize}

\section{Related Work}
\label{sec:related-work}

\paragraph{World Action Models.}
World Action Models combine future video prediction with action generation to learn action-relevant representations from scene evolution.
Early generative policies use predicted frames or visual plans to guide action generation~\cite{unipi,gr1,vpp}, while recent video-action models adapt large-scale video priors through zero-shot, inverse-dynamics, causal, or action-centered formulations~\cite{dreamzero,mimicvideo,lingbotva,bi2026motus,cosmospolicy,gigaworldpolicy}.
Fast-WAM~\cite{yuan2026fastwam} shows that video co-training can improve action learning without inference-time future generation.
X-WAM generates multi-view RGB-D sequences~\cite{guo2026xwam}, whereas GeoSem-WAM and WAM4D introduce auxiliary depth prediction~\cite{ma2026geosemwam,li2026wam4d}.
MECo-WAM adds a training-time 4D expert~\cite{zhang2026mecowam}.
Unlike these methods, \methodabbr{} regularizes video--action representations through pairwise feature relations without training a depth or 4D predictor.

\paragraph{Representation Alignment.}
Aligning intermediate features with frozen foundation models transfers visual and geometric priors to downstream models~\cite{yan2026svam,wang2026vega}.
REPA aligns diffusion features with self-supervised visual features~\cite{yu2025repa,oquab2024dinov2}, while Geometry Forcing and Spatial Forcing use pointwise feature alignment with 3D foundation models for video generation and VLA policies, respectively~\cite{wu2026geometry,wang2025vggt,li2026spatial}.
Relational distillation instead preserves pairwise distances or similarity structures between teacher and student representations~\cite{park2019rkd,tung2019similarity}, with Whitened Similarity Distillation further transferring cross-modal similarity matrices~\cite{huang2023efficient}.
VideoREPA uses pairwise relation alignment to transfer physical priors from foundation models into text-to-video generation~\cite{zhang2025videorepa}.
In contrast, \methodabbr{} aligns intra-frame, cross-view, and action-temporal feature relations in dual-expert WAMs, avoiding direct pointwise correspondence between compact action features and dense geometric targets.

\begin{figure*}[t]
    \centering
    \includegraphics[width=\textwidth]{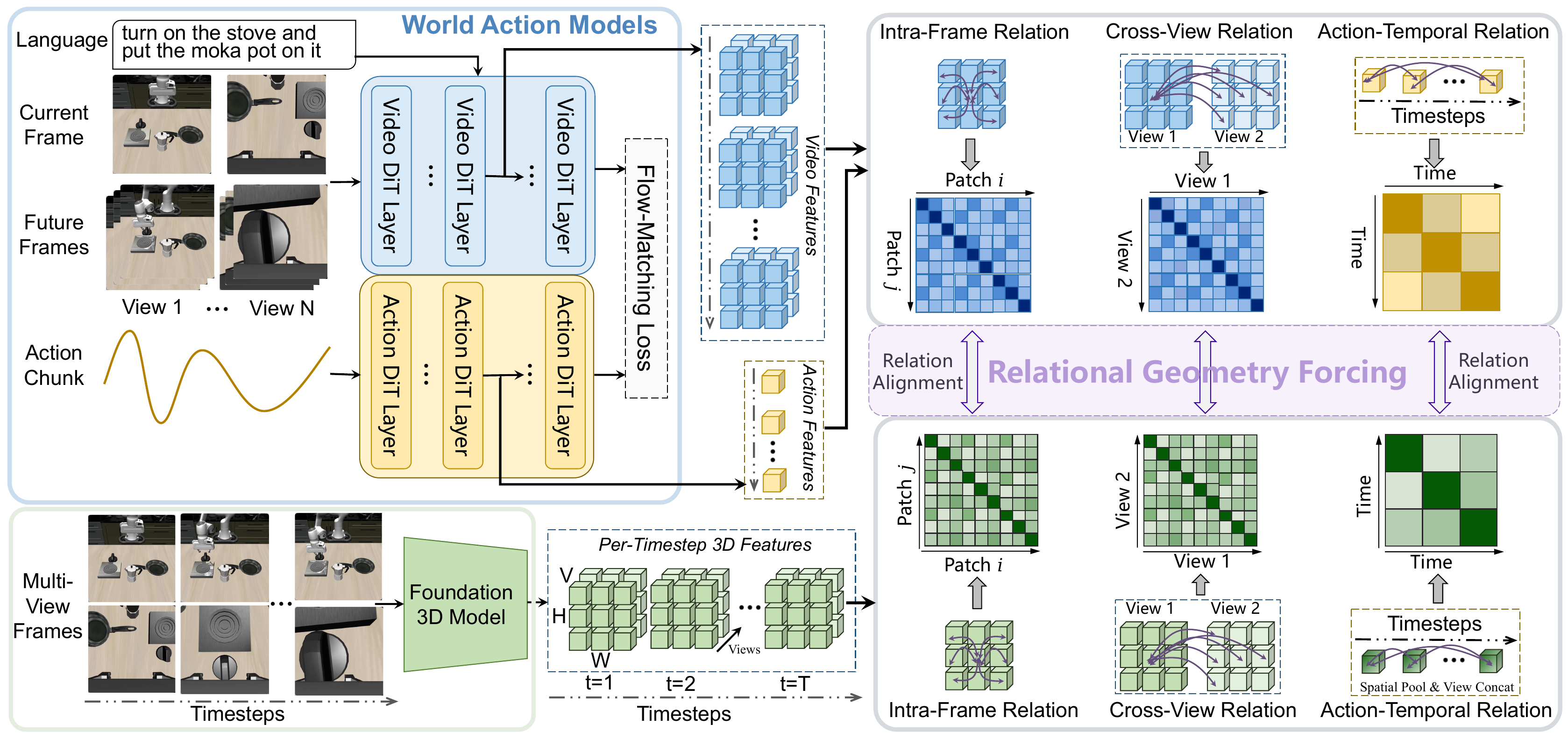}
    \caption{\textbf{Overview of \method{} (\methodabbr{}).}
    The WAM predicts future multi-view frames and an action chunk through its video and action experts, while a frozen 3D foundation model encodes the same frames into per-timestep 3D features.
    \methodabbr{} aligns pairwise relations among intermediate video and action features with those induced by the 3D foundation model through intra-frame, cross-view, and action-temporal objectives.
    All auxiliary components are removed during inference.}
    \label{fig:rgf_method}
\end{figure*}

\paragraph{Geometry-Aware Robot Learning.}
Geometric perception is crucial for spatial reasoning and precise manipulation in robot policies~\cite{ke2024diffuseractor}.
PerAct, RVT, Act3D, and DP3 represent 3D observations with voxels, multi-view features, 3D feature fields, and compact point clouds, respectively~\cite{shridhar2022peract,goyal2023rvt,gervet2023act3d,ze2024dp3}.
Recent VLAs further incorporate point clouds, 3D position encodings, or input-output alignment into language-conditioned control~\cite{li2025pointvla,qu2025spatialvla,li2025bridgevla}.
Alternatively, RoboUniView learns unified-view features via 3D occupancy prediction, while 3D-VLA generates goal images and point clouds for planning~\cite{liu2024robouniview,zhen20243dvla}.
These methods explicitly encode or reconstruct geometry~\cite{han2026gam}.
In contrast, \methodabbr{} uses a frozen 3D foundation model only for training-time supervision and retains the original video--action inference pathway at deployment.

\section{Method}
\label{sec:method}

World Action Models (WAMs) jointly predict future visual observations and executable actions from synchronized multi-view RGB observations and language instructions.
Building on this formulation, \methodabbr{} transfers geometric priors from a frozen 3D foundation model to intermediate features of the video and action experts through pairwise relation alignment.
Figure~\ref{fig:rgf_method} presents the resulting training framework and its three relational objectives.
We begin by formulating WAMs and the geometric teacher in Section~\ref{subsec:preliminaries}, proceed to the relational objectives in Section~\ref{subsec:rgf}, and conclude with the training objective and inference procedure in Section~\ref{subsec:training-inference}.

\subsection{Preliminaries}
\label{subsec:preliminaries}

\paragraph{World Action Models.}
Let $\mathcal{O}_t=\{\mathbf{I}_t^v\}_{v=1}^{V}$ denote a synchronized RGB observation from $V$ camera views at time $t$, let $\mathbf{c}$ denote the language condition, and let $\mathbf{A}_{0:T_a-1}=\{\mathbf{a}_t\}_{t=0}^{T_a-1}$ denote an action chunk.
Conditioned on the current observation and language instruction, a WAM models the joint distribution
\begin{equation}
    p_\theta\left(
    \mathcal{O}_{1:T_v},
    \mathbf{A}_{0:T_a-1}
    \mid
    \mathcal{O}_0,\mathbf{c}
    \right),
\end{equation}
where $\mathcal{O}_{1:T_v}$ contains the future multi-view observations.
We consider a general architecture with a video diffusion transformer (Video DiT) as the video expert and an action diffusion transformer (Action DiT) as the action expert.
The two experts may exchange information through attention or predicted visual representations, and are jointly optimized with video and action flow-matching losses $\mathcal{L}_{\mathrm{FM}}^v$ and $\mathcal{L}_{\mathrm{FM}}^a$, respectively.

We apply geometric supervision to an intermediate layer indexed by $\ell$ in both experts.
Let $\mathbf{H}_{\ell,t,v}^v\in\mathbb{R}^{N_v\times d_v}$ denote the $N_v$ video tokens for view $v$ at time $t$.
We denote the action representation by $\mathbf{H}_\ell^a\in\mathbb{R}^{T_a\times d_a}$, which contains one token $\mathbf{h}_{\ell,t}^a$ for each action step.
This notation does not assume a particular number or arrangement of cameras.

\paragraph{Geometric Teacher.}
We use the frozen Visual Geometry Grounded Transformer (VGGT)~\cite{wang2025vggt} as the geometric teacher.
At each time $t$, VGGT jointly processes the synchronized views $\mathcal{O}_t$ and produces a patch representation
\begin{equation}
    \{\mathbf{P}_t^v\}_{v=1}^{V}
    =
    \Phi_{\mathrm{3D}}(\mathcal{O}_t),
    \qquad
    \mathbf{P}_t^v\in\mathbb{R}^{M_v\times d_g},
\end{equation}
where $M_v$ is the number of teacher patches in view $v$.
The per-view patch features retain spatial resolution and are used for video supervision.
For action supervision, we summarize the scene at each step by spatially pooling every view and concatenating the resulting vectors:
\begin{equation}
    \mathbf{s}_t
    =
    \left[
    \operatorname{Pool}(\mathbf{P}_t^1);
    \ldots;
    \operatorname{Pool}(\mathbf{P}_t^V)
    \right]
    \in\mathbb{R}^{Vd_g},
\end{equation}
VGGT is evaluated independently at each time step while processing all synchronized views jointly.
Its outputs serve only as training targets and are never provided as inputs to the WAM.

\subsection{Relational Geometry Forcing}
\label{subsec:rgf}

\paragraph{Pairwise Feature Relations.}
Pointwise feature alignment assumes that WAM features and their geometric targets encode comparable content in a shared feature space.
This assumption does not hold uniformly in WAMs: video tokens encode local visual regions, whereas action tokens summarize control transitions.
\methodabbr{} therefore matches the pairwise feature relations induced by each representation instead of directly regressing geometric targets.

For a token set $\mathbf{X}=[\mathbf{x}_1,\ldots,\mathbf{x}_N]^\top$, let $\bar{\mathbf{X}}$ denote row-wise $L_2$ normalization.
We define self- and cross-relation operators as
\begin{equation}
    \mathcal{R}(\mathbf{X})
    =
    \bar{\mathbf{X}}\bar{\mathbf{X}}^\top,
    \qquad
    \mathcal{R}(\mathbf{X},\mathbf{Y})
    =
    \bar{\mathbf{X}}\bar{\mathbf{Y}}^\top.
\end{equation}
The entries are pairwise cosine similarities.
Although WAM features and geometric targets may differ in dimensionality, their relation matrices have the same shape once the corresponding token sets are aligned.
We use trainable projection heads $g_{\mathrm{vid}}$ and $g_{\mathrm{act}}$ to decouple the supervised features from the video and action hidden states.
The projected features need not match the dimensionality of the geometric targets.

For a WAM relation matrix $\mathbf{R}^s$ and its geometric target $\mathbf{R}^g$, we compute an element-wise Huber (smooth L1) loss over the valid entries $\Omega$:
\begin{equation}
    \mathcal{D}_{\Omega}(\mathbf{R}^s,\mathbf{R}^g)
    =
    \frac{1}{|\Omega|}
    \sum_{(i,j)\in\Omega}
    \operatorname{Huber}
    \left(R_{ij}^s-R_{ij}^g\right).
\end{equation}
For self-relations, $\Omega$ excludes the diagonal; it also removes entries associated with padded tokens.

\paragraph{Video Relational Alignment.}
We partition the video features by time and camera view and align the VGGT patches with the corresponding WAM feature grid.
When their spatial resolutions differ, the VGGT grid is resampled to obtain $\widetilde{\mathbf{P}}_t^v\in\mathbb{R}^{N_v\times d_g}$.
The intra-frame objective matches the patch relations inside each view:
\begin{equation}
    \mathcal{L}_{\mathrm{intra}}
    =
    \mathbb{E}_{t,v}
    \left[
    \mathcal{D}_{\Omega}
    \left(
    \mathcal{R}(g_{\mathrm{vid}}(\mathbf{H}_{\ell,t,v}^v)),
    \mathcal{R}(\widetilde{\mathbf{P}}_t^v)
    \right)
    \right].
\end{equation}
This objective transfers VGGT's relative organization of spatial regions without direct feature regression.

Intra-frame relations do not capture how different camera projections correspond to the same scene.
For every pair of views $v<u$, we therefore match their cross-relation matrices:
\begin{equation}
\begin{aligned}
    \mathcal{L}_{\mathrm{cross}}
    &=
    \mathbb{E}_{t,\,v<u}
    \Big[
    \mathcal{D}_{\Omega}\Big(\\
    &\quad
    \mathcal{R}\left(
    g_{\mathrm{vid}}(\mathbf{H}_{\ell,t,v}^v),
    g_{\mathrm{vid}}(\mathbf{H}_{\ell,t,u}^v)
    \right),\\
    &\quad
    \mathcal{R}\left(
    \widetilde{\mathbf{P}}_t^v,
    \widetilde{\mathbf{P}}_t^u
    \right)
    \Big)
    \Big].
\end{aligned}
\end{equation}
The resulting matrix has one axis for each view and preserves the cross-view similarity pattern induced by VGGT.

\paragraph{Action Relational Alignment.}
Each action token $\mathbf{h}_{\ell,t}^a$ represents the transition from observation $\mathcal{O}_t$, establishing a temporal correspondence with the geometric scene representation $\mathbf{s}_t$.
However, \methodabbr{} does not regress an action token to this scene vector.
Instead, it compares the step-to-step similarity structure of the action sequence with the temporal similarity structure of the corresponding multi-view scenes.
Let $\mathbf{S}=[\mathbf{s}_0,\ldots,\mathbf{s}_{T_a-1}]^\top$.
The action-temporal objective is
\begin{equation}
    \mathcal{L}_{\mathrm{temporal}}
    =
    \mathcal{D}_{\Omega_a}
    \left(
    \mathcal{R}(g_{\mathrm{act}}(\mathbf{H}_\ell^a)),
    \mathcal{R}(\mathbf{S})
    \right),
\end{equation}
where $\Omega_a$ excludes self-relations and all rows and columns associated with padded action steps.
Thus, the one-to-one temporal correspondence is used to construct aligned sequences, while supervision is applied to relations between pairs of time steps.

\subsection{Training Objective and Inference}
\label{subsec:training-inference}

\paragraph{Overall Objective.}
We combine the three relational terms as
\begin{equation}
    \mathcal{L}_{\mathrm{RGF}}
    =
    w_i\mathcal{L}_{\mathrm{intra}}
    +
    w_c\mathcal{L}_{\mathrm{cross}}
    +
    w_t\mathcal{L}_{\mathrm{temporal}},
\end{equation}
and train the WAM with the full objective
\begin{equation}
    \mathcal{L}
    =
    \mathcal{L}_{\mathrm{FM}}^v
    +
    \mathcal{L}_{\mathrm{FM}}^a
    +
    \lambda_{\mathrm{RGF}}\mathcal{L}_{\mathrm{RGF}},
\end{equation}
where $w_i$, $w_c$, and $w_t$ control the three relational terms and $\lambda_{\mathrm{RGF}}$ balances geometric supervision with the original WAM objectives.

\paragraph{Inference.}
VGGT, the projection heads, and relation computations are used only during training and removed at inference.
The WAM then follows its original video--action pathway using only RGB observations and language instructions, with no geometric inputs or additional inference modules.
Since supervision acts only on intermediate features, \methodabbr{} applies to direct action prediction, joint video--action denoising, and action prediction conditioned on predicted video features.

\section{Experiments}
\label{sec:experiments}
In this section, we structure our experiments to answer the following questions:
\begin{itemize}
    \item How does \model{} compare with existing VLA, WAM, and geometry-aware methods? (Section~\ref{subsec:main-results})
    \item How do the pretrained foundation models and alignment formulation affect geometry transfer across WAM paradigms? (Section~\ref{subsec:geometry-transfer-analysis})
    \item How do the relational objectives contribute individually and jointly? (Section~\ref{subsec:ablations})
    \item How does \methodabbr{} affect learned intra-frame, cross-view, and action-temporal relations? (Section~\ref{subsec:qualitative-analysis})
    \item Does \methodabbr{} consistently improve performance across real-world manipulation tasks? (Section~\ref{subsec:real-world})
\end{itemize}

\subsection{Experimental Setup}
\label{subsec:experimental-setup}

\paragraph{Simulation Benchmarks.}
We evaluate \model{} on LIBERO~\cite{liu2023libero} and RoboTwin 2.0~\cite{chen2025robotwin}, following the training and evaluation protocols of Fast-WAM~\cite{yuan2026fastwam}.
LIBERO includes four suites: Spatial, Object, Goal, and Long.
Each contains 10 tasks and 500 demonstrations.
We evaluate each suite over 500 rollouts with different random seeds, totaling 2,000 rollouts across 40 tasks, and report per-suite and average success rates.

RoboTwin 2.0 includes 50 bimanual manipulation tasks covering diverse object interactions.
For each task, the training set contains 50 demonstrations in the clean setting and 500 demonstrations under scene randomization.
We evaluate 100 rollouts per task in each setting and report the clean, randomized, and average success rates.

\paragraph{Baselines.}
We compare with representative VLA methods, including OpenVLA-OFT~\cite{kim2025openvlaoft}, $\pi_{0.5}$~\cite{intelligence2025pi05}, and Spatial Forcing~\cite{li2026spatial}, as well as WAMs including LingBot-VA~\cite{lingbotva}, Motus~\cite{bi2026motus}, and Fast-WAM~\cite{yuan2026fastwam}.
We further include geometry-aware WAMs X-WAM~\cite{guo2026xwam}, GeoSem-WAM~\cite{ma2026geosemwam}, WAM4D~\cite{li2026wam4d}, and MECo-WAM~\cite{zhang2026mecowam}.

\paragraph{Implementation Details.}
Following Fast-WAM~\cite{yuan2026fastwam}, we build \model{} on Wan2.2-TI2V-5B~\cite{wan2025wan} with a 1B Action DiT and evaluate three representative WAM variants: Uncond, Joint, and IDM, using Uncond by default.
All variants use an action horizon of $H=32$, 9 video frames per training chunk, 10 denoising steps, and CFG scale 1.0 at inference.
Our default \methodabbr{} configuration uses frozen final-layer VGGT-1B features and applies the three relational objectives with equal weights at WAM layer 15, with $\lambda_{\mathrm{RGF}}=0.5$ and Huber loss.

\subsection{Main Results}
\label{subsec:main-results}

\paragraph{LIBERO.}

\begin{table}[t]
\centering
{\small
\setlength{\tabcolsep}{4.pt}
\renewcommand{\arraystretch}{1.0}
\begin{tabularx}{\columnwidth}{l|c|cccc|c}
\toprule
Method & Geo. & Spat. & Obj. & Goal & Long & Avg. \\
\midrule
\multicolumn{7}{l}{\textit{Vision-Language-Action Models}} \\
OpenVLA-OFT & $\times$ & 97.6 & 98.4 & 97.9 & 94.5 & 97.10 \\
$\pi_{0.5}$ & $\times$ & 98.8 & 98.2 & 98.0 & 92.4 & 96.90 \\
Spatial Forcing & $\checkmark$ & 99.4 & 99.6 & 98.8 & 96.0 & 98.50 \\
\midrule
\multicolumn{7}{l}{\textit{World Action Models}} \\
LingBot-VA & $\times$ & 98.5 & 99.6 & 97.2 & 98.5 & 98.50 \\
Motus & $\times$ & 96.8 & 99.8 & 96.6 & 97.6 & 97.70 \\
Fast-WAM & $\times$ & 98.2 & 100.0 & 97.0 & 95.2 & 97.60 \\
GeoSem-WAM & $\checkmark$ & 99.0 & 100.0 & 98.2 & 97.0 & \underline{98.55} \\
MECo-WAM & $\checkmark$ & 98.8 & 100.0 & 98.2 & 95.8 & 98.20 \\
\rowcolor{gray!20}
\textbf{\model{}~(ours)} & $\checkmark$ & 99.4 & 100.0 & 98.6 & 96.6 & \textbf{98.65} \\
\bottomrule
\end{tabularx}
}
\caption{Success rates (\%) on the four LIBERO suites. ``Geo.'' denotes the use of geometric inputs or supervision.}
\label{tab:libero-main}
\end{table}

Table~\ref{tab:libero-main} shows that \model{} achieves the highest average success rate among the compared methods, outperforming both Fast-WAM and existing geometry-aware WAMs.
\methodabbr{} thus strengthens WAM representations during joint video--action training without explicit geometric reconstruction or additional inference modules.
Relative to the Fast-WAM backbone, \model{} yields gains of +1.2, +1.6, and +1.4 on Spatial, Goal, and Long, respectively, while both methods reach 100\% on Object.
The gains across these suites indicate that relational supervision benefits diverse manipulation scenarios.

\paragraph{RoboTwin 2.0.}

\begin{table}[t]
\centering
{\small
\renewcommand{\arraystretch}{1.0}
\begin{tabularx}{\columnwidth}{l|cc|cc|c}
\toprule
Method & P.T. & Geo. & Clean & Rand. & Avg. \\
\midrule
\multicolumn{6}{l}{\textit{Vision-Language-Action Models}} \\
$\pi_{0.5}$ & $\checkmark$ & $\times$ & 82.74 & 76.76 & 79.75 \\
\midrule
\multicolumn{6}{l}{\textit{World Action Models}} \\
Motus & $\checkmark$ & $\times$ & 88.66 & 87.02 & 87.84 \\
LingBot-VA & $\checkmark$ & $\times$ & 92.90 & 91.50 & 92.20 \\
X-WAM & $\checkmark$ & $\checkmark$ & 89.80 & 90.70 & 90.25 \\
WAM4D & $\checkmark$ & $\checkmark$ & \textbf{93.80} & 89.90 & 91.80 \\
Fast-WAM & $\times$ & $\times$ & 91.88 & 91.78 & 91.83 \\
GeoSem-WAM & $\times$ & $\checkmark$ & 92.94 & \underline{92.14} & 92.54 \\
MECo-WAM & $\times$ & $\checkmark$ & \underline{93.26} & 91.98 & \underline{92.62} \\
\rowcolor{gray!20}
\textbf{\model{}~(ours)} & $\times$ & $\checkmark$ & 92.92 & \textbf{92.50} & \textbf{92.71} \\
\bottomrule
\end{tabularx}
}
\caption{Success rates (\%) on RoboTwin 2.0 under clean and randomized evaluation settings. ``P.T.'' denotes embodied-policy pretraining, and ``Geo.'' denotes the use of geometric inputs or supervision.}
\label{tab:robotwin-main}
\end{table}

Table~\ref{tab:robotwin-main} shows that \model{} achieves the highest randomized and average success rates, surpassing MECo-WAM by 0.09 percentage points on average without embodied-policy pretraining.
Relative to Fast-WAM, it improves clean performance from 91.88\% to 92.92\% and randomized performance from 91.78\% to 92.50\%.
\model{} achieves these gains without a depth or 4D predictor, supporting representation regularization under scene variation.

\subsection{Analysis of Geometry Transfer}
\label{subsec:geometry-transfer-analysis}

\begin{table}[t]
\centering
{\small
\renewcommand{\arraystretch}{1.0}
\begin{tabularx}{\columnwidth}{Xcc|cc}
\toprule
\multicolumn{1}{c}{\multirow[c]{2}{*}[-2pt]{\shortstack{Foundation\\Model}}} &
\multicolumn{2}{c|}{Alignment Type} &
\multirow[c]{2}{*}[-2pt]{\shortstack{Average\\SR (\%)}} &
\multirow[c]{2}{*}[-2pt]{$\Delta$} \\
\cmidrule(lr){2-3}
& Video & Action & & \\
\midrule
\multicolumn{5}{l}{\textit{WAM Variant: Direct Action (Uncond)}} \\
\multicolumn{3}{l|}{No alignment} & 97.60 & -- \\
DINOv2 & Pointwise & -- & 98.10 & +0.50 \\
DINOv2 & Relational & Relational & 98.35 & +0.75 \\
VGGT & Pointwise & -- & 97.90 & +0.30 \\
VGGT & Pointwise & Pointwise & 97.55 & $-$0.05 \\
\rowcolor{gray!20}
VGGT & Relational & Relational & \textbf{98.65} & +1.05 \\
\midrule
\multicolumn{5}{l}{\textit{WAM Variant: Joint Generation (Joint)}} \\
\multicolumn{3}{l|}{No alignment} & 98.50 & -- \\
VGGT & Pointwise & -- & 98.75 & +0.25 \\
VGGT & Pointwise & Pointwise & 98.35 & $-$0.15 \\
\rowcolor{gray!20}
VGGT & Relational & Relational & \textbf{99.10} & +0.60 \\
\midrule
\multicolumn{5}{l}{\textit{WAM Variant: Video-then-Action (IDM)}} \\
\multicolumn{3}{l|}{No alignment} & 98.00 & -- \\
VGGT & Pointwise & -- & 98.40 & +0.40 \\
VGGT & Pointwise & Pointwise & 97.95 & $-$0.05 \\
\rowcolor{gray!20}
VGGT & Relational & Relational & \textbf{98.55} & +0.55 \\
\bottomrule
\end{tabularx}
}
\caption{Analysis of geometry transfer on LIBERO. $\Delta$ denotes the gain over the corresponding variant baseline.}
\label{tab:geometry-transfer}
\end{table}

\paragraph{Foundation Model and Alignment.}
Table~\ref{tab:geometry-transfer} shows that pointwise alignment yields only marginal gains when applied to video representations.
Extending it to action features instead drops below the baseline in all three variants, exposing the weakness of pointwise feature alignment.
Relational alignment improves both DINOv2- and VGGT-supervised variants, with VGGT yielding the stronger result.
Notably, relational DINOv2 reaches 98.35\%, exceeding video-only pointwise VGGT at 97.90\%.
This cross-condition comparison suggests that the alignment formulation can matter more than the feature source.
With relational alignment fixed, VGGT further improves over DINOv2 from 98.35\% to 98.65\%, showing that geometric features provide additional benefit under a compatible objective.

\paragraph{Across WAM Variants.}
\methodabbr{} yields gains of +1.05, +0.60, and +0.55 for the Uncond, Joint, and IDM variants, respectively, exceeding pointwise video alignment in each case.
The consistent gains across distinct video--action generation schemes show that \methodabbr{} is compatible with different WAM paradigms.
Together, the DINOv2 and cross-paradigm controls show that the gains are not specific to VGGT or a particular WAM paradigm, supporting the relational formulation as the common source of improvement.

\subsection{Ablation Studies}
\label{subsec:ablations}

\paragraph{Relational Objectives.}

\begin{table}[t]
\centering
{\small
\renewcommand{\arraystretch}{1.0}
\begin{tabular}{ccc|cc}
\toprule
$\mathcal{L}_{\mathrm{intra}}$ & $\mathcal{L}_{\mathrm{cross}}$ & $\mathcal{L}_{\mathrm{temporal}}$ &
Avg. & $\Delta$ \\
\midrule
$\times$ & $\times$ & $\times$ & 97.60 & -- \\
$\checkmark$ & $\times$ & $\times$ & 97.65 & +0.05 \\
$\times$ & $\checkmark$ & $\times$ & 97.70 & +0.10 \\
$\times$ & $\times$ & $\checkmark$ & 97.45 & $-$0.15 \\
$\checkmark$ & $\checkmark$ & $\times$ & 97.90 & +0.30 \\
$\checkmark$ & $\times$ & $\checkmark$ & 97.35 & $-$0.25 \\
$\times$ & $\checkmark$ & $\checkmark$ & 98.00 & +0.40 \\
\rowcolor{gray!20}
$\checkmark$ & $\checkmark$ & $\checkmark$ & \textbf{98.65} & +1.05 \\
\midrule
\multicolumn{5}{l}{\textit{Loss Type (all objectives enabled)}} \\
\multicolumn{3}{l|}{KL divergence} & 98.05 & +0.45 \\
\multicolumn{3}{l|}{Huber (smooth L1)} & \textbf{98.65} & +1.05 \\
\bottomrule
\end{tabular}
}
\caption{Ablation of relational objectives and loss types on LIBERO. $\Delta$ denotes the gain over no alignment.}
\label{tab:terms}
\end{table}

Table~\ref{tab:terms} shows that individual objectives remain close to the baseline, indicating that no relation type is sufficient alone.
Both top-performing two-objective variants include cross-view supervision, highlighting cross-view correspondence as an important geometric cue.
Combining cross-view with action-temporal supervision yields +0.40, whereas combining intra-frame with action-temporal supervision yields $-$0.25, further showing that cross-view correspondence stabilizes temporal supervision.
Although action-temporal supervision alone yields $-$0.15, adding it to intra-frame and cross-view supervision raises the gain from +0.30 to +1.05.
This reversal suggests that action-temporal relations become more useful when jointly grounded by intra-frame and cross-view relations.
The full model reaches +1.05, compared with +0.40 for the best two-objective variant, indicating non-additive complementarity among the three objectives.

\paragraph{Loss Function.}

KL divergence normalizes relation scores into probability distributions, discarding their absolute scale.
Huber instead matches each relation value directly and uses a linear penalty for large residuals, preserving magnitude while limiting outlier influence.
Huber's 98.65\% result versus 98.05\% for KL supports retaining relation magnitudes in the supervision signal.

\subsection{Qualitative Analysis}
\label{subsec:qualitative-analysis}

\begin{figure}[t]
    \centering
    \includegraphics[width=\columnwidth]{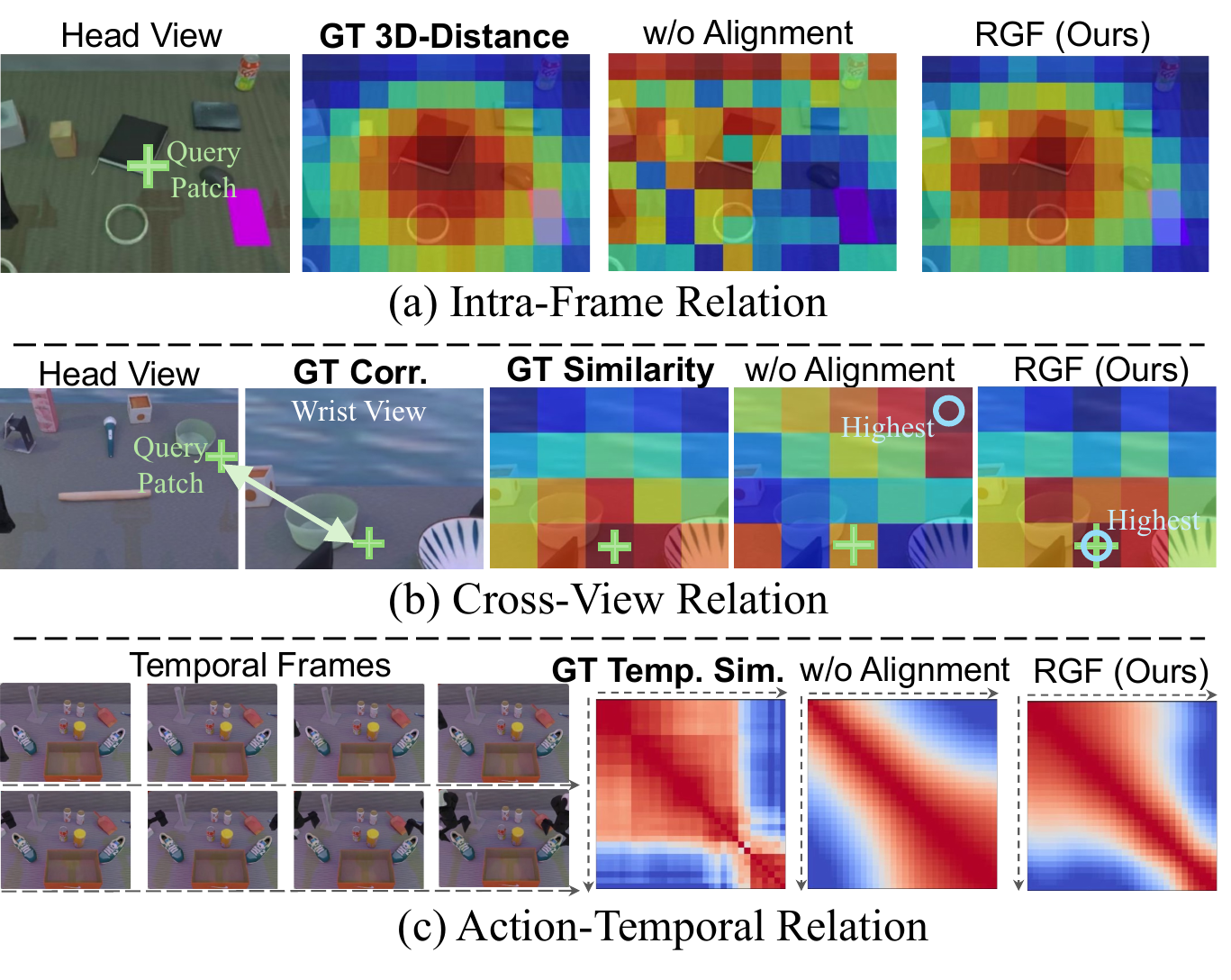}
    \caption{\textbf{Intra-frame, cross-view, and action-temporal relations learned by \methodabbr{}.}
    (a) Intra-frame similarity maps for the marked query patch, with the ground-truth 3D distance map as reference.
    (b) Cross-view similarity maps in the wrist view.
    The green plus marks the ground-truth correspondence, and the circle marks the highest-similarity patch.
    (c) Action-temporal similarity matrices for a rollout with nonuniform motion, compared with the scene-evolution reference.
    Across all three settings, \methodabbr{} follows the geometric reference more closely than the unaligned WAM.}
    \label{fig:relation-visualization}
\end{figure}

\paragraph{Intra-Frame and Cross-View Relations.}
We examine whether \methodabbr{} makes intermediate video features reflect scene geometry by comparing their similarity maps with geometric references in Figure~\ref{fig:relation-visualization}.
The intra-frame and cross-view references are derived from simulator 3D geometry.
Each similarity map reports the cosine similarity between the query feature and all candidate features.
In Figure~\ref{fig:relation-visualization}(a), the unaligned map is fragmented and deviates from the ground-truth 3D distance pattern, whereas \methodabbr{} recovers a smoother similarity structure around the query patch.
In Figure~\ref{fig:relation-visualization}(b), the highest-similarity patch from \methodabbr{} lies near the ground-truth correspondence across the head and wrist views, while the unaligned WAM selects a distant region.
Together, these results show that \methodabbr{} captures both within-view geometric proximity and cross-view correspondence.

\begin{figure*}[t]
    \centering
    \includegraphics[width=\textwidth]{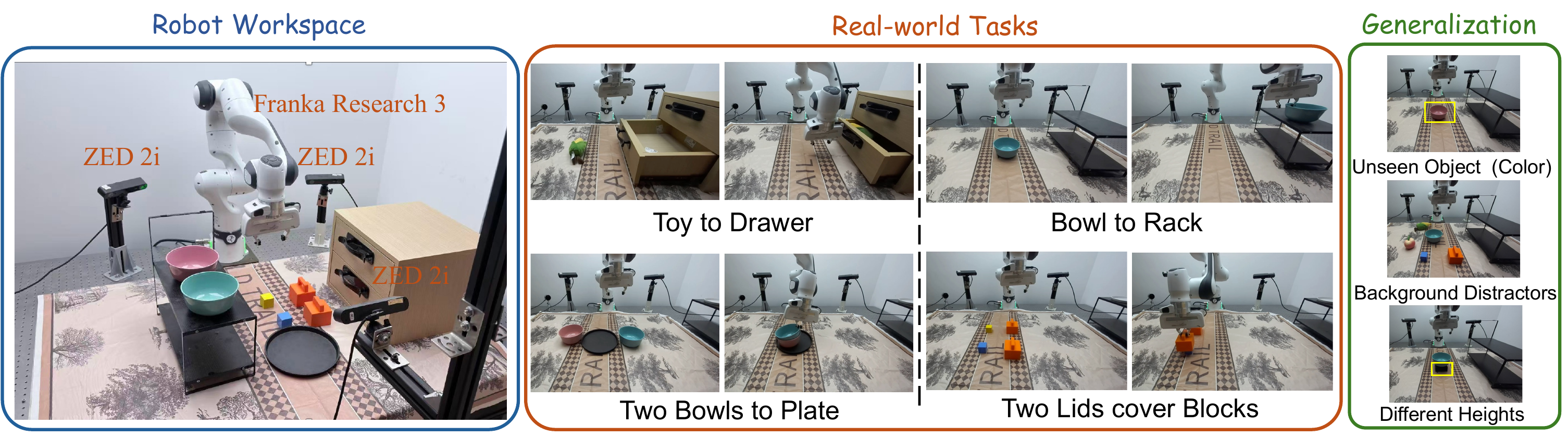}
    \caption{\textbf{Real-world setup, tasks, and generalization settings.}
    Left: a Franka Research 3 arm observed by three ZED 2i cameras.
    Middle: the four evaluation tasks.
    Right: Bowl-to-Rack generalization settings with an unseen object color, background distractors, and altered placement heights.}
    \label{fig:real-world-setup}
\end{figure*}

\begin{table*}[!t]
\centering
{\small
\setlength{\tabcolsep}{4.5pt}
\renewcommand{\arraystretch}{1.0}
\begin{tabular}{lcccccccc}
\toprule
\multirow[c]{2}{*}[-2pt]{Method} &
\multicolumn{4}{c}{Basic Tasks} &
\multicolumn{3}{c}{Generalization} &
\multirow[c]{2}{*}[-2pt]{\shortstack{Avg.\\Succ. (\%)}} \\
\cmidrule(lr){2-5}\cmidrule(lr){6-8}
& Toy$\to$Drawer & Bowl$\to$Rack & Bowls$\to$Plate & Lids$\to$Blocks
& Unseen Object & Distractors & Height & \\
\midrule
$\pi_{0.5}$ & \textbf{65} & 70 & 65 & 50 & 60 & 30 & 40 & 54.29 \\
Fast-WAM & 55 & \textbf{85} & 60 & 40 & 65 & 40 & 55 & 57.14 \\
\rowcolor{gray!20}
\textbf{\model{}~(ours)} & 60 & \textbf{85} & \textbf{70} & \textbf{55} & \textbf{70} & \textbf{45} & \textbf{75} & \textbf{65.71} \\
\bottomrule
\end{tabular}
}
\caption{Real-world success rates (\%) over 20 rollouts per setting. Generalization settings are Bowl-to-Rack variants with an unseen object color, background distractors, or altered placement heights. Best results per column are in bold.}
\label{tab:real-world}
\end{table*}

\paragraph{Action-Temporal Relations.}
For action-temporal analysis, we compare the action feature relation matrix with the scene-evolution reference.
The rollout in Figure~\ref{fig:relation-visualization}(c) starts slowly and accelerates, so early frames remain similar over a longer interval than later frames.
The scene-evolution reference reflects this motion as a wide band that progressively narrows.
The unaligned WAM produces a nearly uniform-width band, whereas \methodabbr{} recovers the nonuniform pattern.
\methodabbr{} therefore aligns action-temporal relations with the changing rate of geometric scene evolution.

\subsection{Real-World Experiments}
\label{subsec:real-world}

\paragraph{Robot Setup.}
As shown in Figure~\ref{fig:real-world-setup}, we evaluate real-world manipulation on a Franka Research 3 arm observed by three ZED 2i cameras.
Each policy predicts 32-step chunks of end-effector (EEF) actions.

\paragraph{Tasks and Protocol.}
The four basic tasks cover single-object transfer, placement onto a rack, and two-stage manipulation of multiple objects.
For each task, we collect 50 demonstrations and fine-tune all methods on the same task data.
We evaluate each method over 20 rollouts per setting with randomized initial object placements shared across methods.
To assess generalization, we construct three Bowl-to-Rack variants with an unseen object color, background distractors, or altered placement heights, none of which appear during training.

\paragraph{Results.}
Table~\ref{tab:real-world} shows that \model{} attains the highest average success rate, outperforming Fast-WAM by 8.6 percentage points and achieving the best or tied-best result in six of seven settings.
On the basic tasks, the largest gains over Fast-WAM occur on \emph{Two Bowls to Plate} and \emph{Two Lids cover Blocks}, with absolute gains of +10 and +15, respectively.
Both tasks require coordinating multiple objects over successive manipulation stages, suggesting that relational supervision becomes more beneficial as spatial and temporal dependencies increase.
On the original Bowl-to-Rack task, \model{} and Fast-WAM both achieve 85\%, providing a matched baseline for the three generalization settings.
Under unseen color, distractor, and height variations, \model{} improves over Fast-WAM by +5, +5, and +20, respectively.
The larger gain under height variation suggests that relational supervision is particularly useful when the distribution shift changes spatial configuration.

\section{Conclusion}
\label{sec:conclusion}
In this paper, we investigated how to transfer geometric priors to World Action Models without explicit geometric reconstruction.
We identified a representation mismatch in pointwise alignment between dense geometric targets and compact action features, and introduced \method{} (\methodabbr{}) to transfer geometry through pairwise feature relations without direct feature regression.
\methodabbr{} combines intra-frame, cross-view, and action-temporal objectives to supervise video and action features while preserving the original video--action inference pathway.
Experiments in simulation and on a real Franka robot demonstrate competitive manipulation performance, real-world generalization, and consistent gains across diverse WAM paradigms.
These results indicate that relational alignment provides a compatible interface for geometric supervision across heterogeneous WAM features.
The current formulation focuses on pairwise cosine relations, leaving explicit metric scale and contact structure unexplored.
Future work may incorporate these richer geometric cues and longer-horizon dynamics while retaining \methodabbr{}'s training-time relational supervision.

\bibliography{aaai2027}

\clearpage
\appendix
\section*{Appendix}
\numberwithin{equation}{section}
\setcounter{table}{0}
\renewcommand{\thetable}{A\arabic{table}}
\setcounter{figure}{0}
\renewcommand{\thefigure}{A\arabic{figure}}
\setcounter{algorithm}{0}
\renewcommand{\thealgorithm}{A\arabic{algorithm}}

This appendix provides additional details and results for the main text.
Appendix~\ref{sec:supp-method} expands the pointwise alignment baseline, relational objectives, and training procedure.
Appendix~\ref{sec:supp-implementation} documents the model architecture, training hyperparameters, and compute usage.
Appendix~\ref{sec:supp-libero} reports suite-level LIBERO results.
Appendix~\ref{sec:supp-robotwin} provides task-level RoboTwin 2.0 results under clean and randomized settings.
Appendix~\ref{sec:supp-ablations} presents ablations of the alignment layer and $\lambda_{\mathrm{RGF}}$.
Appendix~\ref{sec:supp-real-world} visualizes complete real-world rollouts for the four basic tasks.

\section{Method Details}
\label{sec:supp-method}
This section complements the method description in the main paper.
We first formalize the pointwise alignment, then describe the practical construction of the relational objectives and the training workflow.

\subsection{Pointwise Alignment}
\label{subsec:supp-pointwise}
We implement pointwise feature alignment as a controlled alternative to \methodabbr{}, following the representation-alignment principle of REPA~\cite{yu2025repa}.
Let
\begin{equation}
    d_{\mathrm{cos}}(\mathbf{x},\mathbf{y})
    =
    1-
    \frac{\mathbf{x}^{\top}\mathbf{y}}
    {\lVert\mathbf{x}\rVert_2\lVert\mathbf{y}\rVert_2}
\end{equation}
denote cosine distance.
For video alignment, the VGGT patch grid is bilinearly resampled to the corresponding WAM feature grid.
The pointwise video objective is
\begin{equation}
    \mathcal{L}_{\mathrm{pw}}^v
    =
    \mathbb{E}_{t,v}
    \left[
    \frac{1}{N_v}
    \sum_{i=1}^{N_v}
    d_{\mathrm{cos}}
    \left(
    g_{\mathrm{vid}}^{\mathrm{pw}}
    (\mathbf{h}_{\ell,t,v,i}^v),
    \widetilde{\mathbf{p}}_{t,v,i}
    \right)
    \right],
    \label{eq:supp-pointwise-video}
\end{equation}
where $\widetilde{\mathbf{p}}_{t,v,i}$ is the geometric target at the same spatial location.

For action alignment, we use only spatially pooled patch features, excluding VGGT camera tokens.
Each action step $t$ is paired with the observation at the same step, and the geometric target for view $v$ is
$\mathbf{q}_t^v=\operatorname{Pool}(\mathbf{P}_t^v)$.
The action projector output is reshaped to $V\times d_g$, yielding one vector per view:
\begin{equation}
    \mathcal{L}_{\mathrm{pw}}^a
    =
    \mathbb{E}_{t}
    \left[
    \frac{1}{V}
    \sum_{v=1}^{V}
    d_{\mathrm{cos}}
    \left(
    \left[
    g_{\mathrm{act}}^{\mathrm{pw}}
    (\mathbf{h}_{\ell,t}^a)
    \right]_v,
    \mathbf{q}_t^v
    \right)
    \right].
    \label{eq:supp-pointwise-action}
\end{equation}
When enabled, the video and action pointwise terms each use a loss weight of $0.5$.
The video-only control disables $\mathcal{L}_{\mathrm{pw}}^a$ while keeping all other settings unchanged.
For the video-only DINOv2 pointwise control, final-layer ViT-L patch features replace VGGT features, and the video projector output dimension is adjusted accordingly~\cite{oquab2024dinov2}.

\subsection{Relational Objectives}
\label{subsec:supp-relations}
We extract final-layer features from a frozen VGGT and cache them before WAM training~\cite{wang2025vggt}.
For video supervision, cached VGGT features are selected at RGB frames corresponding to the VAE latent timesteps, and their patch grids are bilinearly resampled to the WAM grid.
The projected WAM features and geometric targets are normalized along the feature dimension before constructing the full cosine relation matrices.

For intra-frame relations, we compute all feature pairs within each camera view and remove self-relations on the diagonal.
For cross-view relations, we enumerate every unique view pair $v<u$ and retain all cross-view entries.
For action-temporal supervision, the 32-step action sequence is paired with 32 pooled scene descriptors obtained from the corresponding consecutive observations.
This construction is independent of the nine-frame video subsampling and therefore requires no temporal interpolation.
Rows and columns associated with padded action steps are masked, and action self-relations are removed.
For an entry-wise residual $e$, we use the smooth L1 (Huber) penalty
\begin{equation}
    \rho_{\beta}(e)
    =
    \begin{cases}
        \dfrac{e^2}{2\beta}, & |e|<\beta,\\[3pt]
        |e|-\dfrac{\beta}{2}, & \text{otherwise},
    \end{cases}
    \qquad \beta=1.
    \label{eq:supp-smooth-l1}
\end{equation}
The masked relation loss is
\begin{equation}
    \mathcal{D}_{\Omega}(\mathbf{R}^s,\mathbf{R}^g)
    =
    \frac{1}{|\Omega|}
    \sum_{(i,j)\in\Omega}
    \rho_{1}\!\left(R_{ij}^s-R_{ij}^g\right),
    \label{eq:supp-masked-relation}
\end{equation}
where $\Omega$ contains the valid entries defined above.
The loss is averaged over valid entries, views, timesteps, and samples.
The intra-frame, cross-view, and action-temporal terms are equally weighted, and the overall RGF weight is $\lambda_{\mathrm{RGF}}=0.5$.

\subsection{Training Procedure}
\label{subsec:supp-training-procedure}
Algorithm~\ref{alg:rgf-training} summarizes RGF training.
VGGT features are precomputed, and only the WAM and projection heads receive gradients.

\begin{algorithm}[t]
\caption{Training with Relational Geometry Forcing}
\label{alg:rgf-training}
\begin{algorithmic}[1]
\REQUIRE WAM $f_{\theta}$, projection heads $g_{\mathrm{vid}}$ and $g_{\mathrm{act}}$, cached VGGT targets
\FOR{each optimization step}
    \STATE Sample a synchronized video--action batch and diffusion timesteps
    \STATE Run $f_{\theta}$ and extract layer-15 video and action features
    \STATE Load the cached VGGT targets for the sampled window
    \STATE Construct intra-frame, cross-view, and action-temporal relations
    \STATE Compute the RGF and video/action flow-matching losses
    \STATE Update $f_{\theta}$, $g_{\mathrm{vid}}$, and $g_{\mathrm{act}}$
\ENDFOR
\end{algorithmic}
\end{algorithm}

RGF is enabled throughout fine-tuning without a separate warm-up stage.
At inference, the cached targets, projection heads, and relation computations are removed, so the deployed WAM follows its original video--action pathway.

\section{Implementation Details}
\label{sec:supp-implementation}
This section documents the model configurations and training settings used in all experiments.

\subsection{Model Architecture}
\label{subsec:supp-architecture}
All models are initialized from the pretrained Wan2.2-TI2V-5B video diffusion model~\cite{wan2025wan}.
The pretrained video DiT serves as the video expert, while the action expert is initialized by linearly interpolating the video-DiT weights to the smaller hidden dimension.
Their configurations are summarized in Table~\ref{tab:supp-architecture}.

\begin{table}[t]
\centering
\small
\begin{tabular}{lcc}
\toprule
Configuration & Video Expert & Action Expert \\
\midrule
Transformer layers & 30 & 30 \\
Hidden dimension & 3072 & 1024 \\
FFN dimension & 14336 & 4096 \\
Attention heads & 24 & 24 \\
\bottomrule
\end{tabular}
\caption{Shared architecture of the video and action experts.}
\label{tab:supp-architecture}
\end{table}

RGF uses separate two-layer MLP projection heads with a SiLU activation.
The video and action heads map $3072\!\rightarrow\!2048$ and $1024\!\rightarrow\!2048$, respectively.
For the DINOv2 control, the video projection dimension is changed to 1024 to match the ViT-L features.
All projection heads are removed at inference.

We evaluate three WAM paradigms under this shared backbone~\cite{yuan2026fastwam}.
The \emph{Uncond} paradigm predicts actions directly from the current visual context, while retaining video co-training.
The \emph{Joint} paradigm jointly denoises future video and action representations.
The \emph{IDM} paradigm first predicts future video representations and then conditions action prediction on them.

\subsection{Training Hyperparameters}
\label{subsec:supp-optimization}
We use AdamW with a learning rate of $1\times10^{-4}$, weight decay of $10^{-2}$, and a cosine schedule following a linear warm-up over the first 5\% of training.
Training uses bf16 mixed precision, a per-device batch size of 16, and gradient clipping at 1.0.
LIBERO and RoboTwin 2.0 models are trained for 10 and 5 epochs, respectively.
At inference, all models use 10 denoising steps and a classifier-free guidance scale of 1.0.

\subsection{Compute Usage}
\label{subsec:supp-compute}
LIBERO experiments are trained on 8 NVIDIA H20 GPUs and require approximately 16 hours per run.
RoboTwin 2.0 experiments use 32 NVIDIA H20 GPUs and require approximately 65 hours per run.

\section{Detailed LIBERO Results}
\label{sec:supp-libero}
This section provides suite-level results that complement the aggregate analyses on LIBERO~\cite{liu2023libero} in the main paper.

\subsection{Geometry Transfer by Suite}
\label{subsec:supp-libero-transfer}
\begin{table*}[t]
\centering
{\small
\setlength{\tabcolsep}{4.5pt}
\renewcommand{\arraystretch}{1.0}
\begin{tabular}{lcc|cccccc}
\toprule
Foundation Model & Video & Action & Spatial & Object & Goal & Long & Avg. & $\Delta$ \\
\midrule
\multicolumn{9}{l}{\textit{WAM Paradigm: Direct Action (Uncond)}} \\
\multicolumn{3}{l|}{No alignment} & 98.2 & 100.0 & 97.0 & 95.2 & 97.60 & -- \\
DINOv2 & Pointwise & -- & 99.2 & 100.0 & 97.8 & 95.4 & 98.10 & +0.50 \\
DINOv2 & Relational & Relational & 98.4 & 100.0 & 99.0 & 96.0 & 98.35 & +0.75 \\
VGGT & Pointwise & -- & 98.4 & 99.8 & 98.0 & 95.4 & 97.90 & +0.30 \\
VGGT & Pointwise & Pointwise & 99.2 & 98.6 & 98.2 & 94.2 & 97.55 & $-$0.05 \\
\rowcolor{gray!20}
VGGT & Relational & Relational & 99.4 & 100.0 & 98.6 & 96.6 & \textbf{98.65} & +1.05 \\
\midrule
\multicolumn{9}{l}{\textit{WAM Paradigm: Joint Generation (Joint)}} \\
\multicolumn{3}{l|}{No alignment} & 99.6 & 99.4 & 98.2 & 96.8 & 98.50 & -- \\
VGGT & Pointwise & -- & 99.8 & 99.8 & 98.6 & 96.8 & 98.75 & +0.25 \\
VGGT & Pointwise & Pointwise & 99.4 & 99.6 & 98.4 & 96.0 & 98.35 & $-$0.15 \\
\rowcolor{gray!20}
VGGT & Relational & Relational & 99.8 & 100.0 & 98.8 & 97.8 & \textbf{99.10} & +0.60 \\
\midrule
\multicolumn{9}{l}{\textit{WAM Paradigm: Video-then-Action (IDM)}} \\
\multicolumn{3}{l|}{No alignment} & 98.8 & 97.8 & 97.8 & 97.6 & 98.00 & -- \\
VGGT & Pointwise & -- & 99.4 & 98.0 & 98.2 & 98.0 & 98.40 & +0.40 \\
VGGT & Pointwise & Pointwise & 98.8 & 97.8 & 98.0 & 97.2 & 97.95 & $-$0.05 \\
\rowcolor{gray!20}
VGGT & Relational & Relational & 99.4 & 98.8 & 97.6 & 98.4 & \textbf{98.55} & +0.55 \\
\bottomrule
\end{tabular}
}
\caption{Suite-level success rates (\%) for geometry-transfer controls on LIBERO.
Each suite uses 500 rollouts, following the training and evaluation protocol of the main paper.
$\Delta$ denotes the gain over the corresponding paradigm baseline.}
\label{tab:supp-libero-transfer}
\end{table*}

Table~\ref{tab:supp-libero-transfer} reports suite-level geometry-transfer results for all three WAM paradigms.
For the Uncond paradigm, extending VGGT pointwise alignment to action features reduces the average from 97.90\% to 97.55\%, with the largest drops on Object and Long.
Relational alignment instead reaches 98.65\% and improves Goal and Long over the unaligned baseline.
Relational DINOv2 also reaches 98.35\%, exceeding video-only pointwise VGGT and confirming that the alignment formulation is important beyond the foundation-model choice.

The same trend holds across WAM paradigms.
Full relational alignment improves the Joint average from 98.50\% to 99.10\%, with gains on all four suites.
For IDM, it improves Spatial, Object, and Long while remaining close on Goal, yielding a net gain from 98.00\% to 98.55\%.

\subsection{Objective Ablations by Suite}
\label{subsec:supp-libero-objectives}
\begin{table*}[t]
\centering
{\small
\setlength{\tabcolsep}{5pt}
\renewcommand{\arraystretch}{1.0}
\begin{tabular}{ccc|cccccc}
\toprule
$\mathcal{L}_{\mathrm{intra}}$ &
$\mathcal{L}_{\mathrm{cross}}$ &
$\mathcal{L}_{\mathrm{temporal}}$ &
Spatial & Object & Goal & Long & Avg. & $\Delta$ \\
\midrule
$\times$ & $\times$ & $\times$ & 98.2 & 100.0 & 97.0 & 95.2 & 97.60 & -- \\
$\checkmark$ & $\times$ & $\times$ & 98.0 & 99.8 & 98.6 & 94.2 & 97.65 & +0.05 \\
$\times$ & $\checkmark$ & $\times$ & 98.4 & 99.6 & 97.6 & 95.2 & 97.70 & +0.10 \\
$\times$ & $\times$ & $\checkmark$ & 98.8 & 99.0 & 98.2 & 93.8 & 97.45 & $-$0.15 \\
$\checkmark$ & $\checkmark$ & $\times$ & 97.8 & 100.0 & 97.8 & 96.0 & 97.90 & +0.30 \\
$\checkmark$ & $\times$ & $\checkmark$ & 97.2 & 99.6 & 98.2 & 94.4 & 97.35 & $-$0.25 \\
$\times$ & $\checkmark$ & $\checkmark$ & 98.8 & 100.0 & 97.6 & 95.6 & 98.00 & +0.40 \\
\rowcolor{gray!20}
$\checkmark$ & $\checkmark$ & $\checkmark$ & 99.4 & 100.0 & 98.6 & 96.6 & \textbf{98.65} & +1.05 \\
\midrule
\multicolumn{9}{l}{\textit{Loss Type (all objectives enabled)}} \\
\multicolumn{3}{l|}{KL divergence} & 99.8 & 100.0 & 97.2 & 95.2 & 98.05 & +0.45 \\
\rowcolor{gray!20}
\multicolumn{3}{l|}{Huber (smooth L1)} & 99.4 & 100.0 & 98.6 & 96.6 & \textbf{98.65} & +1.05 \\
\bottomrule
\end{tabular}
}
\caption{Suite-level ablation of the three relational objectives and loss types on LIBERO.
Each suite uses 500 rollouts, and $\Delta$ denotes the gain over no alignment.}
\label{tab:supp-libero-objectives}
\end{table*}

Table~\ref{tab:supp-libero-objectives} reports the suite-level objective and loss ablations.
Individual objectives remain close to the baseline and provide no substantial overall gain, although their effects differ by suite.
Intra-frame and action-temporal supervision improve Goal but reduce Long, while cross-view supervision provides the most stable single-objective result.
Among two-objective configurations, cross-view with action-temporal supervision performs best, reaching 98.00\%.
Combining all three objectives produces the strongest and most balanced result, improving Spatial, Goal, and Long while retaining 100\% on Object.

Replacing Huber with KL divergence raises Spatial from 99.4\% to 99.8\% but reduces Goal and Long by 1.4 points each.
The lower 98.05\% average indicates that directly preserving relation magnitudes with Huber provides a better balance across suites.

\section{Detailed RoboTwin 2.0 Results}
\label{sec:supp-robotwin}
This section first introduces the compared baselines and then analyzes the task-level results in Table~\ref{tab:supp-robotwin-detail} on RoboTwin 2.0~\cite{chen2025robotwin}.

\begin{table*}[p]
\centering
\small
\setlength{\tabcolsep}{1.pt}
\resizebox{0.98\textwidth}{!}{%
\begin{tabular}{@{}l*{16}{c}@{}}
\toprule
Task & \multicolumn{2}{c}{$\pi_{0.5}$} & \multicolumn{2}{c}{Motus} & \multicolumn{2}{c}{LingBot} & \multicolumn{2}{c}{Fast-WAM} & \multicolumn{2}{c}{X-WAM} & \multicolumn{2}{c}{GeoSem} & \multicolumn{2}{c}{WAM4D} & \multicolumn{2}{c}{\textbf{RGF (Ours)}} \\
\cmidrule(lr){2-3}\cmidrule(lr){4-5}\cmidrule(lr){6-7}\cmidrule(lr){8-9}\cmidrule(lr){10-11}\cmidrule(lr){12-13}\cmidrule(lr){14-15}\cmidrule(lr){16-17}
    & Clean & Rand. & Clean & Rand. & Clean & Rand. & Clean & Rand. & Clean & Rand. & Clean & Rand. & Clean & Rand. & Clean & Rand. \\
\midrule
Adjust Bottle & 100 & 99 & 89 & 93 & 90 & 94 & 100 & 100 & 100 & 99 & 100 & 100 & 100 & 99 & 100 & 99 \\
Beat Block Hammer & 96 & 93 & 95 & 88 & 96 & 98 & 99 & 97 & 98 & 96 & 98 & 98 & 99 & 97 & 100 & 97 \\
Blocks Ranking RGB & 92 & 85 & 99 & 97 & 99 & 98 & 100 & 100 & 99 & 95 & 100 & 98 & 100 & 97 & 100 & 99 \\
Blocks Ranking Size & 49 & 26 & 75 & 63 & 94 & 96 & 94 & 98 & 76 & 82 & 91 & 96 & 98 & 84 & 92 & 94 \\
Click Alarmclock & 98 & 89 & 100 & 100 & 99 & 100 & 100 & 100 & 98 & 99 & 100 & 100 & 100 & 100 & 100 & 100 \\
Click Bell & 99 & 66 & 100 & 100 & 100 & 100 & 100 & 100 & 100 & 100 & 100 & 100 & 100 & 100 & 100 & 100 \\
Dump Bin Bigbin & 92 & 97 & 95 & 91 & 89 & 96 & 97 & 96 & 90 & 96 & 96 & 95 & 92 & 94 & 99 & 98 \\
Grab Roller & 100 & 100 & 100 & 100 & 100 & 100 & 100 & 100 & 100 & 100 & 100 & 100 & 100 & 100 & 100 & 100 \\
Handover Block & 66 & 57 & 86 & 73 & 99 & 78 & 95 & 81 & 88 & 79 & 96 & 81 & 96 & 93 & 98 & 88 \\
Handover Mic & 98 & 97 & 78 & 63 & 94 & 96 & 99 & 100 & 88 & 89 & 100 & 100 & 94 & 94 & 100 & 100 \\
Hanging Mug & 18 & 17 & 38 & 38 & 40 & 28 & 58 & 62 & 46 & 55 & 72 & 67 & 64 & 57 & 100 & 100 \\
Lift Pot & 96 & 85 & 96 & 99 & 100 & 99 & 100 & 100 & 100 & 99 & 100 & 100 & 100 & 88 & 88 & 95 \\
Move Can Pot & 51 & 55 & 34 & 74 & 94 & 97 & 90 & 88 & 80 & 84 & 90 & 95 & 96 & 97 & 100 & 100 \\
Move Pillbottle Pad & 84 & 61 & 93 & 96 & 99 & 99 & 100 & 99 & 96 & 98 & 99 & 98 & 100 & 99 & 83 & 75 \\
Move Playingcard Away & 96 & 84 & 100 & 96 & 100 & 99 & 100 & 100 & 100 & 98 & 100 & 100 & 100 & 100 & 66 & 61 \\
Move Stapler Pad & 56 & 42 & 83 & 85 & 91 & 79 & 77 & 64 & 67 & 70 & 73 & 63 & 79 & 76 & 99 & 100 \\
Open Laptop & 90 & 96 & 95 & 91 & 92 & 94 & 98 & 100 & 96 & 97 & 99 & 100 & 100 & 66 & 70 & 50 \\
Open Microwave & 34 & 77 & 95 & 91 & 82 & 86 & 62 & 45 & 89 & 92 & 75 & 50 & 61 & 65 & 84 & 88 \\
Pick Diverse Bottles & 81 & 71 & 90 & 91 & 89 & 82 & 80 & 85 & 91 & 92 & 87 & 86 & 98 & 92 & 99 & 96 \\
Pick Dual Bottles & 93 & 63 & 96 & 90 & 100 & 99 & 100 & 96 & 99 & 100 & 100 & 97 & 100 & 99 & 95 & 96 \\
Place A2B Left & 87 & 82 & 88 & 79 & 97 & 93 & 95 & 93 & 90 & 87 & 93 & 95 & 97 & 86 & 90 & 93 \\
Place A2B Right & 87 & 84 & 91 & 87 & 97 & 95 & 93 & 99 & 92 & 89 & 95 & 95 & 94 & 79 & 96 & 96 \\
Place Bread Basket & 77 & 64 & 91 & 94 & 97 & 95 & 91 & 93 & 90 & 91 & 90 & 92 & 93 & 97 & 95 & 96 \\
Place Bread Skillet & 85 & 66 & 86 & 83 & 95 & 90 & 90 & 93 & 90 & 96 & 91 & 98 & 93 & 87 & 72 & 73 \\
Place Burger Fries & 94 & 87 & 98 & 98 & 97 & 95 & 96 & 99 & 97 & 99 & 96 & 100 & 96 & 97 & 100 & 100 \\
Place Can Basket & 62 & 62 & 81 & 76 & 81 & 84 & 71 & 69 & 84 & 82 & 70 & 70 & 87 & 83 & 99 & 99 \\
Place Cans Plasticbox & 94 & 84 & 98 & 94 & 100 & 99 & 99 & 96 & 99 & 98 & 99 & 99 & 100 & 100 & 91 & 94 \\
Place Container Plate & 99 & 95 & 98 & 99 & 99 & 97 & 96 & 100 & 98 & 100 & 98 & 98 & 98 & 91 & 100 & 100 \\
Place Dual Shoes & 75 & 75 & 93 & 87 & 94 & 89 & 94 & 88 & 83 & 81 & 93 & 91 & 96 & 97 & 96 & 97 \\
Place Empty Cup & 100 & 99 & 99 & 98 & 100 & 100 & 100 & 100 & 98 & 99 & 100 & 100 & 100 & 100 & 98 & 99 \\
Place Fan & 87 & 85 & 91 & 87 & 99 & 93 & 96 & 96 & 84 & 92 & 97 & 96 & 100 & 95 & 89 & 87 \\
Place Mouse Pad & 60 & 39 & 66 & 68 & 93 & 96 & 83 & 89 & 84 & 86 & 89 & 89 & 96 & 97 & 86 & 84 \\
Place Object Basket & 80 & 76 & 81 & 87 & 91 & 88 & 89 & 88 & 85 & 87 & 89 & 85 & 88 & 90 & 93 & 94 \\
Place Object Scale & 86 & 80 & 88 & 85 & 96 & 95 & 90 & 97 & 93 & 89 & 92 & 92 & 99 & 97 & 91 & 95 \\
Place Object Stand & 91 & 85 & 98 & 97 & 99 & 96 & 90 & 94 & 97 & 96 & 91 & 91 & 98 & 100 & 99 & 98 \\
Place Phone Stand & 81 & 81 & 87 & 86 & 97 & 97 & 97 & 99 & 75 & 80 & 98 & 99 & 94 & 81 & 100 & 99 \\
Place Shoe & 92 & 93 & 99 & 97 & 98 & 98 & 96 & 99 & 97 & 99 & 97 & 99 & 100 & 99 & 96 & 98 \\
Press Stapler & 87 & 83 & 93 & 98 & 85 & 82 & 90 & 97 & 94 & 90 & 94 & 96 & 97 & 94 & 93 & 95 \\
Put Bottles Dustbin & 84 & 79 & 81 & 79 & 87 & 91 & 95 & 90 & 85 & 95 & 92 & 88 & 87 & 85 & 91 & 90 \\
Put Object Cabinet & 80 & 79 & 88 & 71 & 85 & 87 & 94 & 89 & 66 & 76 & 92 & 87 & 81 & 53 & 91 & 86 \\
Rotate QRcode & 89 & 87 & 89 & 73 & 96 & 91 & 93 & 89 & 84 & 83 & 96 & 95 & 95 & 95 & 88 & 87 \\
Scan Object & 72 & 65 & 67 & 66 & 96 & 91 & 89 & 92 & 86 & 79 & 89 & 89 & 92 & 68 & 88 & 86 \\
Shake Bottle & 99 & 97 & 100 & 97 & 100 & 97 & 100 & 100 & 99 & 99 & 100 & 100 & 100 & 99 & 100 & 100 \\
Shake Bottle Horizontally & 99 & 99 & 100 & 98 & 100 & 99 & 100 & 100 & 100 & 99 & 100 & 100 & 100 & 99 & 100 & 100 \\
Stack Blocks Three & 91 & 76 & 91 & 95 & 99 & 98 & 95 & 97 & 97 & 95 & 96 & 98 & 98 & 97 & 96 & 95 \\
Stack Blocks Two & 97 & 100 & 100 & 98 & 100 & 98 & 100 & 100 & 100 & 100 & 100 & 100 & 100 & 100 & 100 & 99 \\
Stack Bowls Three & 77 & 71 & 79 & 87 & 86 & 83 & 80 & 81 & 88 & 82 & 91 & 83 & 83 & 84 & 82 & 85 \\
Stack Bowls Two & 95 & 96 & 98 & 98 & 94 & 98 & 92 & 98 & 98 & 98 & 93 & 98 & 96 & 95 & 98 & 98 \\
Stamp Seal & 79 & 55 & 93 & 92 & 96 & 97 & 90 & 94 & 93 & 95 & 92 & 95 & 96 & 87 & 91 & 92 \\
Turn Switch & 62 & 54 & 84 & 78 & 44 & 45 & 61 & 59 & 61 & 72 & 58 & 65 & 60 & 64 & 64 & 74 \\
\midrule
\textbf{Average} & 82.74 & 76.76 & 88.66 & 87.02 & 92.90 & 91.50 & 91.88 & 91.78 & 89.80 & 90.70 & 92.94 & 92.14 & 93.80 & 89.90 & 92.92 & 92.50 \\
\bottomrule
\end{tabular}%
}
\caption{Per-task success rates (\%) on RoboTwin 2.0 under clean and randomized evaluation settings. The average row follows the aggregate values reported in the main paper.}
\label{tab:supp-robotwin-detail}
\end{table*}

\subsection{Compared Baselines}
\label{subsec:supp-robotwin-baselines}
The compared methods span three representative policy families.
$\pi_{0.5}$ is a large pretrained VLA that directly generates robot actions~\cite{intelligence2025pi05}.
Motus and LingBot-VA are pretrained WAMs that couple visual dynamics with action generation~\cite{bi2026motus,lingbotva}, whereas Fast-WAM is the non-pretrained direct-action WAM that serves as our backbone~\cite{yuan2026fastwam}.
X-WAM is a pretrained 4D WAM that jointly predicts multi-view RGB-D videos and actions~\cite{guo2026xwam}.
GeoSem-WAM introduces auxiliary geometry and semantic prediction during training without embodied-policy pretraining~\cite{ma2026geosemwam}.
WAM4D initializes from pretrained LingBot-VA and transfers depth priors through training-time spatial registers~\cite{li2026wam4d}.
RGF-WAM shares the Fast-WAM backbone and uses relational supervision without explicit geometry prediction.
MECo-WAM is omitted because a complete 50-task breakdown is not publicly available.
Table~\ref{tab:supp-robotwin-detail} reports the clean and randomized success rates for all 50 tasks.

\subsection{Task-Level Analysis}
\label{subsec:supp-robotwin-analysis}
Table~\ref{tab:supp-robotwin-detail} shows that RGF-WAM improves the Fast-WAM average from 91.88\% to 92.92\% in the clean setting and from 91.78\% to 92.50\% under randomization.
Its clean average is comparable to GeoSem-WAM (92.94\%) and remains close to WAM4D (93.80\%), while its randomized average is the highest among the compared methods.
Across the 50 tasks, RGF-WAM achieves the best or tied-best result on 23 clean tasks and 18 randomized tasks.
The improvements span short-horizon object interactions, fine-grained contact operations, and longer multi-stage manipulation, indicating that relational supervision is not restricted to a single task category or horizon.
Performance nevertheless remains uneven across individual tasks.
This variation suggests that geometric relations improve robustness and spatial reasoning but do not replace task-specific action modeling.

\section{Extra Ablation Studies}
\label{sec:supp-ablations}
This section studies the sensitivity of RGF to the supervised WAM layer and the overall relational loss weight in Tables~\ref{tab:supp-layer-ablation} and~\ref{tab:supp-lambda-ablation}.
Unless stated otherwise, all configurations use the Uncond WAM paradigm, VGGT supervision, all three relational objectives, and Huber loss.

\subsection{Alignment Layer}
\label{subsec:supp-layer-ablation}
\begin{table}[t]
\centering
\small
\begin{tabular}{lccccc}
\toprule
Layer & Spatial & Object & Goal & Long & Avg. \\
\midrule
No alignment & 98.2 & 100.0 & 97.0 & 95.2 & 97.60 \\
L8 & 98.6 & 99.8 & 97.8 & 95.6 & 97.95 \\
\rowcolor{gray!20}
L15 & 99.4 & 100.0 & 98.6 & 96.6 & \textbf{98.65} \\
L22 & 99.0 & 100.0 & 98.4 & 95.4 & 98.20 \\
\bottomrule
\end{tabular}%
\caption{Ablation of the supervised WAM layer on LIBERO. L15 is the default setting.}
\label{tab:supp-layer-ablation}
\end{table}

Table~\ref{tab:supp-layer-ablation} shows that the middle layer L15 provides the strongest average result.
Compared with L8 and L22, L15 performs consistently better on Spatial, Goal, and Long while retaining 100\% on Object.
Aligning the shallower L8 still improves over the unaligned baseline, but the injected geometric relations must pass through many subsequent transformer layers and may be gradually reshaped.
At the deeper L22, the features are more specialized toward the original video and action denoising objectives, which can reduce their compatibility with external geometric supervision.
The middle layer L15 therefore provides a better balance between preserving spatial structure and encoding task-relevant dynamics.

\subsection{RGF Loss Weight}
\label{subsec:supp-lambda-ablation}
\begin{table}[t]
\centering
\small
\begin{tabular}{lccccc}
\toprule
$\lambda_{\mathrm{RGF}}$ & Spatial & Object & Goal & Long & Avg. \\
\midrule
0 (off) & 98.2 & 100.0 & 97.0 & 95.2 & 97.60 \\
0.25 & 98.4 & 100.0 & 98.8 & 96.4 & 98.40 \\
\rowcolor{gray!20}
0.5 & 99.4 & 100.0 & 98.6 & 96.6 & \textbf{98.65} \\
1.0 & 98.0 & 100.0 & 98.0 & 94.2 & 97.55 \\
\bottomrule
\end{tabular}%
\caption{Ablation of the RGF loss weight on LIBERO. $\lambda_{\mathrm{RGF}}=0.5$ is the default setting.}
\label{tab:supp-lambda-ablation}
\end{table}

As shown in Table~\ref{tab:supp-lambda-ablation}, a weight of 0.25 already improves over the unaligned baseline, indicating that a relatively weak relational constraint is beneficial.
Increasing $\lambda_{\mathrm{RGF}}$ to 0.5 further improves the average from 98.40\% to 98.65\%, with gains on Spatial and Long.
In contrast, a larger weight of 1.0 reduces the average to 97.55\% and produces the largest degradation on Long, suggesting that excessive relational supervision can interfere with the original video and action objectives.
We therefore use 0.5 as the default balance between geometric regularization and WAM training.

\section{Real-World Rollout Visualizations}
\label{sec:supp-real-world}
\begin{figure*}[!t]
    \centering
    \includegraphics[width=\textwidth]{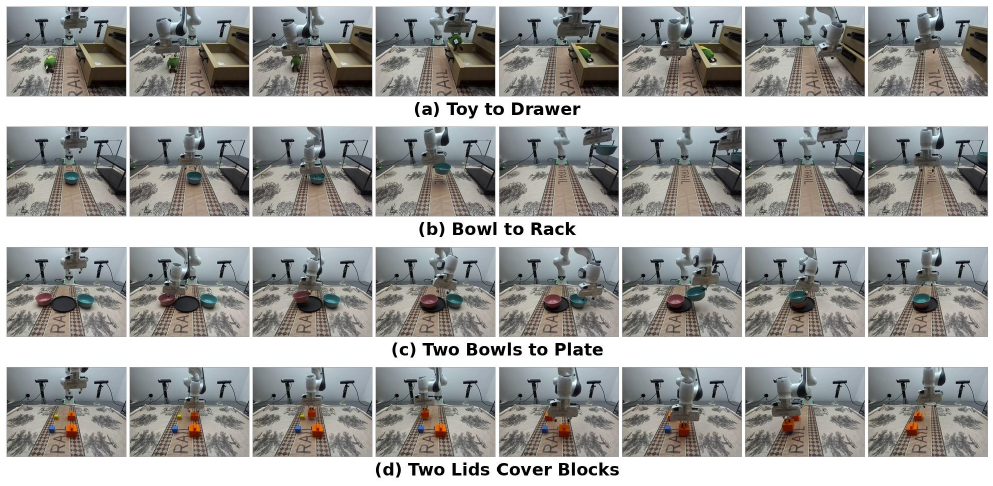}
    \caption{\textbf{Complete real-world task executions.}
    Each row shows eight uniformly sampled frames from one successful rollout.
    From top to bottom: \emph{Toy to Drawer}, \emph{Bowl to Rack}, \emph{Two Bowls to Plate}, and \emph{Two Lids Cover Blocks}.}
    \label{fig:supp-real-rollouts}
    \end{figure*}

Figure~\ref{fig:supp-real-rollouts} visualizes representative successful executions of the four basic real-world tasks.
Each row contains eight frames uniformly sampled from a complete rollout and ordered from left to right.
The examples cover single-object transfer, precise placement, and multi-stage manipulation involving multiple objects.

\FloatBarrier

\end{document}